\documentclass[10pt, a4paper]{article}
\usepackage{lrec}
\usepackage{graphicx}
\usepackage{tabularx}
\usepackage{soul}

\usepackage{algorithm}
\usepackage[noend]{algpseudocode}

\usepackage{booktabs}
\usepackage{epstopdf}
\usepackage[latin1]{inputenc}
\usepackage{multicol}
\usepackage{hyperref}
\usepackage{xstring}

\title{A Tweet-based Dataset for Company-Level Stock Return Prediction}

\name{Karolina Sowinska and Pranava Madhyastha}

\address{Department of Computing, Imperial College London \\
         180 Queen's Gate, Kensington, London SW7 2AZ\\
          \{karolina.sowinska18, pranava\}@imperial.ac.uk\\
         }

\abstract{
Public opinion influences events, especially related to stock market movement, in which a subtle hint can influence the local outcome of the market. 
In this paper, we present a dataset that allows for company-level analysis of tweet based impact on one-, two-, three-, and seven-day stock returns. 
Our dataset consists of $862,231$ labelled instances from twitter in English, we also release a cleaned subset of $85,176$ labelled instances to the community.
We also provide baselines using standard machine learning algorithms and a multi-view learning-based approach that makes use of different types of features. Our dataset, scripts and models are publicly available at: \url{https://github.com/ImperialNLP/stockreturnpred}. \\ 
\newline 
}
\begin{document}

\maketitleabstract
\section{Introduction}
 Traditionally, investors used to build their stock return prediction models based exclusively on numerical data. They would then refer to qualitative information like news articles to finalise their conviction about a given investment idea. However, this process can be hugely intuitive and error-prone. 
 \cite{lo2004adaptive} suggests that we cannot process vast amounts of data or even make rational decisions based only on simple analytical analysis. The persistent inherent human biases could be eliminated from the investment process by the means of objective machine learning algorithms. Although psychological biases can be thought of as weaknesses of human
thought, they can also be viewed as revealing something profound about the complexity of human cognition. Investors' decisions are based on a rich understanding of the companies they
are investing in, the broader environment those companies operate in (including competitors, governments, regulators, etc.), and an understanding of the factors that influence the market behaviour. While capturing the entirety is a tall task, in this paper we tackle a pragmatic, plausible yet difficult task that hinges on building models that processes the abundance of qualitative data from social media posts. This is especially useful for building models that exploit textual information to predict upward or downward movement in the stock price over the following week. This model has a practical application for a buy-and-hold investor. Before the investor executes a trade, the investor would want to make sure that he will buy the stock at the lowest possible price. Thus, understanding if the stock price of a particular company will increase or decrease over the next couple of days is cost-effective, especially when dealing with large transaction sizes.

 Despite the numerous research efforts to predict the stock market returns with Twitter posts, prices of individual companies are typically not taken into account. The most popular studies in the field investigate the stock market movement as a whole \cite{Bollen,Li2014,Oliveira2017}. The problem with such an approach is that aggregate models are not useful for portfolio managers who are interested in predicting the stock price of a particular company.

 In this paper, we present a dataset which allows for processing the abundance of qualitative data from social media posts to investigate whether the textual information predicts the upward or downward movement in the stock price over one-, two-, three- and seven-day period. 
 


\section{NLP for Stock Return Prediction}
The advancements in Natural Language Processing have dramatically shaped queries on stock return prediction in recent years. Social media content and news articles reflect the moods and opinions that people have about certain companies, which in turn can be a predictor of the company's stock returns. Such emotions can be described as a sentiment - a positive sentiment value captures messages with an overall positive tone, and negative sentiment value captures messages with an overall negative tone. The sentiment of textual data such as news articles or tweets can be detected and administered to investigations of the stock returns, often as a sole input to a model. 
One prominent study that made an attempt at this was conducted in recent work by \cite{li2017discovering}. The study developed a novel way called Social Media Data Analyzer--Sentiment Analysis (SMeDA-SA) for mining ambiguous temporal data, such as that contained in the Twitter posts. SMeDA-SA is a pipeline comprised of: 
\begin{enumerate}
    \item  NLP techniques are applied to classify a tweet's sentiment into five categories (Positive+, Positive, Neutral, Negative, Negative-)
     \item  Target concepts that the user is interested in (e.g., a particular company, such as ``Apple Inc.") are identified.
      \item  The method makes use of an association discover algorithm to discover all concepts that are associated with the target concept (e.g., the ``MacBook Pro" can be discovered to be associated with ``Apple Inc.")
       \item  The method uses an algorithm developed for semantic analysis to generate a concept map for the target concept (in our example, ``Apple Inc") based on other concepts that are found to be associated with it
        \item The degree of association is determined between the concept maps and the five sentiment categories
        \item Finally, another algorithm is applied to see if there is a correlation between stock prices and the sentiment detected in the tweets.
\end{enumerate}

Previous work has also looked for the association between the sentiment measure and the daily movement of the stock price. The strength of this study is that it recognises the possible delay that the social media content might have on the stock price. Since the sentiment value may not be immediately reflected in the company's stock price, the study investigates multiple lags. The results generally show that 
for the 30 selected companies, the association between sentiment and the stock price seems to be strongest when the lag is set to T+3. This is an interesting finding which stands in opposition to the Efficient Market Hypothesis \cite{malkiel1989efficient} which assumes that all publicly available information is immediately reflected in the stock price. To evaluate the effectiveness of the proposed approach and accuracy of stock movement prediction based solely on social media data, the authors performed experiments on 30 NASDAQ and NYSE listed companies. For mining social media data for public sentiment, they collected approximately 15 million records of Twitter data that mentioned these 30 companies either directly or indirectly by mentioning their products or services.

Another relevant study conducted by \cite{Bollen} uses Twitter sentiment analysis for stock return prediction. The authors study the relationship between daily sentiment of a collection of Tweet posts and Dow Jones Industrial Average (DJIA)\footnote{https://www.marketwatch.com/investing/index/djia}, which is an index indicating the overall movement of the stock market prices. The authors obtain sentiment specific information followed by
a Granger causality analysis \cite{seth2015granger} in which they correlate DJIA values to two sentiment time series obtained from the sentiment tools. They also deploy a Self-Organising Fuzzy Neural Network model to test the hypothesis that the prediction accuracy of DJIA prediction models can be improved by including measurements of public mood. Forecasting accuracy is measured in terms of the average
Mean Absolute Percentage Error (MAPE) and the direction accuracy (up or down). The study finds that the ``calm" mood is the most predictive of the direction of the stock market, with the 86.7\% direction accuracy. Other investigated moods, namely ``happy", ``sure", ``vital", ``kind", ``alert", do not have a similar predictive power. The study further substantiates the idea that public mood (in this case the calmness of the public) can be predictive of the stock market movement, even as tested with simple, third-party, general sentiment analysis tools. Importantly it reveals the clear relationship between the social media sentiment and stock returns. 
A few studies have also employed multi-view learning to the sentiment analysis problem. These can be categorized into three groups:
a) co-training;
b) multiple kernel learning; and
c) subspace learning.
Co-training style algorithms train alternately to maximise the consensus on two distinct views of the data, multiple kernel learning algorithms combine a few kernels either linearly or non-linearly, whereas subspace learning algorithms aim to obtain a latent subspace shared by multiple views by assuming that the input views are generated from the latent subspace \cite{Xu2013}. The first and second approach seem to be more frequently incorporated for the stock return prediction task. For example, \cite{Li2014} combines the sentiment from market news and historical stock prices in order to improve the prediction accuracy of stock returns. The authors employ multiple kernel learning in order to combine the two views, and their results outperform three baseline methods (which use only one information source or use a naive combination of the two sources) on test data. Similarly, \cite{Shynkevich2016} use the same type of multi-view learning algorithm to combine information extracted from multiple news categories for the prediction of healthcare stocks. Integrating the information from five different kernels gives superior results to single-view models in this study. The usage of co-training algorithms for the stock return prediction was noted in the study conducted by \cite{Ben-Ami}. The authors first label the tweets into positive, neutral, and negative categories with a text-based Sentiment Analysis system. This system is assumed to classify tweets into positive and negative categories with satisfactory precision, but insufficient recall. Thus, the tweets that fall into the neutral category cannot be assumed to contain only neutral sentiment. Multi-view learning is employed to solve this issue. The authors recognised that they can utilise the meta-data of a tweet, as long as it is conditionally independent of the tweet's content given its polarity. Using this additional representation, the study trains a binary SVM with positive and negative tweets as the training data. The classifier then produces the sentiment for the neutral set of tweets. The results confirm again that, when multiple sources are combined, they produce much better accuracy than single-view data. 
There exists a considerable body of literature on applying NLP approaches to stock return prediction problem. In general, previous work shows that sentiment analysis, as well as fusing heterogeneous information sources, can improve the results achieved with solely numerical inputs.

\section{The Dataset} 
Our dataset comprises of two parts, the twitter (social media) based textual information as samples and the stock return information as labels. In the following section, we discuss the process of obtaining both sets of information. We begin by discussing the accumulation of financial information.
\subsection{Financial data}
We investigate tweets that mention one of the top-$100$ stocks of public companies that own the most valuable brands, according to their ranking in \textit{Financial Times Top 100 Global Brands 2019}. We choose to investigate such companies because we believe they elicit significantly stronger emotional responses in people, which is reflected in the sentiment of social media posts. Choosing to investigate stocks which constitute a specific share index (e.g. S\&P 500) could result in less meaningful results because, despite their large market capitalisation, they can be less well known to the public. This is often the case with the parent companies with multiple divisions. For example, Whitbread PLC is not a household name, and we assume that it is unlikely to be tweeted about. Its subsidiaries - Costa Coffee and Premier Inn hotels are much more popular, but tweets relating to those names would not be captured when using the index approach. Thus, we believe that investigating valuable brands instead can lead to more meaningful results. Some brands featuring in the ranking are owned by non-public companies, i.e. their shares are not traded on any stock exchange. An example of such a brand is Louis Vuitton, which is owned by Louis Vuitton Malletier, a private company. Such companies were not taken into account, because naturally, they do not have a public share price. 
For each of the 100 stocks we collect the following financial data:
\begin{enumerate}
\item    \textbf{Last Price} - Closing stock price
\item    \textbf{Volume}  -  The amount of stock that was traded on a given day  
\item    \textbf{10 Day Volatility}  -  The range of price change a stock experienced over the last $10$ days
\item    \textbf{30 Day Volatility}  - The range of price change a stock experienced over the last $30$ days
\end{enumerate}
 
We collect the data with daily frequency. It is typical of financial data to contain missing values for Saturdays and Sundays because the stock market is then closed. Thus, we decide to use the option of imputing the weekend values with the preceding Fridays' values. Financial metrics which might be predictive of a short term stock return are the traded volume of the stock and its volatility. These are indicators commonly used in trading, which is buying and selling stocks on very short time horizons \cite{Zhou2018,Oliveira2017}. Additional metrics could be collected in the future research in order to increase the prediction accuracy, e.g. moving averages, turnover (the total value of the trade shares) and a more granular data on the prices (e.g. opening price, maximum price, minimum price).
\subsection{Twitter data}
\subsubsection{Collection}
 We used $~2$TB of unprocessed Tweets, and filter them using our custom scripts to create the desired dataset. The data was provided in the `Spritzer' version\footnote{https://gist.github.com/jemerick/126173}, the lightest and shallow of Twitter grabs, which captures a 1\% random sample of public tweets. We collect tweets from between July and October 2018. The reuse of Twitter data is permitted by Terms of Service, as well as it is within the Privacy Policy \cite{ahmed2017using}.
 \subsubsection{Filtering}
 In order to create a structured dataset of about $1$ million relevant tweets, we filter nearly $~2$B of raw data. We obtain the data in a compressed format, and we use the $glob$ module to find pathnames matching a specified pattern. The json files with tweets are then decompressed using $shutil$ module. The json files have one-minute granularity. One month of raw data contains roughly $43,200$ json files, each having a size of about 10Mb, amounting to over $400$GB in total. We adopt a simple approach of manually splitting the data into multiple batches, whereby creating daily data frames. In this process, we filter the tweets by language using a ``len" meta-data tag attached to each tweet. Moreover, in the process of creating the data frames, we discard most of the meta-data tags associated with each tweet, leaving only two fields: text and date of creation. This is to comply with the personal data regulations, as well as to reduce the size of the dataset. After the dataset is reduced to contain only English tweets, we perform a keyword search to retain the tweets which mention at least one of the $100$ chosen companies. We release our scripts and codebase publically: \url{https://github.com/ImperialNLP/stockreturnpred}. Choosing the appropriate keywords is paramount to reducing noise in the dataset. We propose two sets of keywords:
Common company name, e.g. Santander v/s
Official Twitter company username, e.g. ``@bancosantander"
We believe that investigating company Twitter usernames produces a more relevant dataset. This is because this approach reduces the risk of erroneously including keywords which disguise as a company name. For example, using a keyword ``Ford" captures tweets not only about the American automaker but also about the sexual assault allegation made by Christine Ford against Brett Kavanaugh in July 2018. Capturing only tweets containing ``@Ford" username eliminates this noise. However, excluding common company names reduces the dataset almost tenfold. Thus, we decide to include tweets selected by either of the two sets of keywords. As a result, we release two versions of the dataset.
In order to process the vast amount of data efficiently, we design Algorithm \ref{keywordsearch} which traverses each batch of data only once.

\begin{algorithm}
  \caption{Filter tweets by keywords}\label{keywordsearch}
  \begin{algorithmic}[1]
      \State{stocks \Comment{A list of 100 companies' names/usernames}}
      \State{tweetList  \Comment{A list of all tweets}}
      \State{keywords = [ ] }
      
      \For{tweet in tweetList}
      \State{findKeywords(tweet, keywods)}
      
      \EndFor
      
    \Procedure{findKeywords}{tweet,keywods}

    \State{tokenized = tokenize(tweet)}
  
      \For{stock in stocks}
      \If{stock in tokenized}
      \State{keywords.append(stock)}
            \State{return keywords}
        \EndIf
        \EndFor
    \State{keywords.append('NaN')}
    \State{return keywords}
    \EndProcedure
  \end{algorithmic}
\end{algorithm}

Tweets containing no keywords are removed from the batch. After the keyword search is performed on all batches, we join the reduced data frames to create the final Twitter dataset, which contains 862,231 tweets.
\subsection{Joining the two datasets}
The two datasets need to be merged together on common indices. Firstly, we look at a tweet's keyword and creation date. Next, we look up the keyword in the dictionary to find the column index that the given company has in the financial dataset. Having located the relevant column, we search a time series of values to find the one relating to the tweet's date of creation. In order for the dates in both dataset to match, they are turned from strings into $datatime$ objects, and their formats are normalised. 
\subsection{Grouping samples into independent events}
The resulting dataset is not a typical time series, as there is no guarantee that each company is mentioned in tweets at least once a day. Due to this fragmented nature of the data, we assume that imputing missing value would introduce too much noise. Instead, we make an assumption that every day for any company is an independent event. In the context of the sentiment analysis, we believe that this assumption is likely to hold true. For example, a swathe of angry customers tweeting about their bad experience with a given company on a given day might have an influence on the company's next day stock price. However, this does not mean that the company's fundamentals have changed. Thus, we hypothesise that it is likely due to the chance that these customers felt a certain way about a given company on a given day. Tweets published next day mentioning the same company might be referring to something unrelated to the past. We suggest that researchers and practitioners create a group for every independent event. An independent event is formed of all tweets relating to a given company on a given day. If we eliminate groups that contain less than 10 tweets, we obtain a total of 5352 groups from the initial 862,231 tweets.

\begin{table}[h]
\caption{Example groups of independent events}
\centering

\label{tab:grouping}
\begin{tabular}{lll}
 Date & Keyword & Group ID \\ \hline
02/03/2018 & Apple & 0 \\
02/03/2018 & Apple & 0 \\
02/03/2018 & Apple & 0 \\
05/03/2018 & Apple & 1 \\
02/03/2018 & IBM & 2 \\
02/03/2018 & IBM & 2 \\
05/03/2018 & IBM & 3 \\
07/03/2018 & Starbucks & 4
\end{tabular}
\end{table}
It follows that we treat all tweets belonging to a given group as one sample.
\subsection{Attaching labels}
We decide to attach four labels to our data: one-, two-, three-, and seven-day stock returns. A stock return is a percentage difference between a company's stock price on the tweet's creation date and the future price. Selecting the aforementioned time lags for generating labels is in alignment with commonly investigated lags in literature \cite{Bollen}, \cite{Li2014}.
Further, if predicting the direction of the stock price rather than an absolute magnitude is of interest, we recommend transforming the returns into 3 categories: negative return (below 0\%), no change (0\%) and positive return (above 0\%). The financial dataset imputes weekend price values with Friday closing prices. Thus, there are a lot of ``no change" labels just due to the fact that a tweet was published on Friday. In order to eliminate this noise, we advise excluding the ``no change" label. The resulting dataset will be ready for a binary classification task.

\section{Analysis of the Dataset}
In this section, we describe the basic analysis of the dataset. This includes an investigation into vocabulary, the general theme in the collected data and further an analysis of the financial data.
\subsection{Vocabulary Analysis}
With our $862,231$, we have $142,088$ distinct tokens in our vocabulary. In a first glance, we observed that among top $20$ unigrams there
are words such as ``dr", ``christine", ``blasey", ``kavanaugh" and ``ford". The reason for their prevalence is that filtering the tweets by the company keyword ``Ford", we capture all the tweets relating to Dr Christine Blasey Ford's sexual assault allegations against Brett Kavanaugh in September 2018\footnote{\url{https://en.wikipedia.org/wiki/Christine_Blasey_Ford}}. This story dominated the keyword `Ford' for a significant proportion of time during our collection. To maintain a real-world scenario, we maintain this version of the dataset as a noisy dataset. We then remove all tweets with ``Ford" keyword, as we believe that a vast majority of them does not relate to the actual
automaker. After removing the ``Ford" keywords. This results in much more familiar twitter vocabulary distribution with unigrams and bigrams being consistent with a random subsample of English twitter samples.

\begin{table}[!h]
    \centering
    \begin{tabular}{lrrr}
    \toprule
     \textbf{Company}    & \textbf{Topic 1} & \textbf{Topic 2} & \textbf{Topic 3} \\
     \bottomrule
     Adobe     &  Photoshop & Illustrator & graphic \\
     Apple & iPhone & MacBook & Music \\
     Audi & R8 & car & new \\
     Aviva & fund & insurance & community \\
     Bank of America & Account & Bank & Money \\
     Disney & movie & channel & love \\
     Expedia & hotel & night & star \\
     Heineken & lagos & drinking & advertisemen \\
     IBM & watson & cloud & blockchain \\
     McDonalds & burger & fries & man\\
     \bottomrule
    \end{tabular}
    \caption{Selected companies and topics}
    \label{tab:companies}
\end{table}
\begin{table*}[]
    \centering
    \begin{tabular}{lccccccc}
    \toprule
         & \textbf{1-return} & \textbf{2-return} & \textbf{3-return} & \textbf{7-return} & \textbf{Volume} & \textbf{Volatality10D} & \textbf{Voalatility30D}\\
         \midrule
    mean &  0.000961 & 0.001654 & 0.001414 & 0.003201 & 1.312400e+07 & 25.469685 & 25.802760 \\
    std & 0.019171 & 0.023578 & 0.027333 & 0.043026 & 1.585069e+07 & 19.411511 & 13.970805 \\
    min & -0.146650 & -0.173554 & -0.177851 & -0.204959 & 1.000000e+00 & 0.619000 & 4.525000 \\
    max & 0.233973 & 0.243639 & 0.243639 & 0.267113 & 3.148332e+08 & 124.137000 & 87.685000\\
    \bottomrule
    \end{tabular}
    \caption{Descriptive statistics of financial data}
    \label{tab:returns}
\end{table*}
\subsection{Topic Analysis}
To analyse the distribution of distinct themes present in the data, we perform a topic analysis of the dataset. We use both Latent Dirichlet Allocation~\cite{blei2003latent} and topic clustering based on Non-negative Matrix Factorization~\cite{liu2013multi}. In general, we found that it was difficult to get a coherent picture of the entire dataset. However, topic modelling with each of our companies gave us much more coherent topics. In Table~\ref{tab:companies}, we tabulate a selection of companies and their associated topics. We see that in general, our data contains relevant information about the companies.
The theme analysis performed by both topic modelling and text clustering reveals that the
most commonly discussed subject on Twitter is videos, music, iPhone, sport, Trump, and
shopping. This is an expected outcome given the nature of social media. After performing
the topic modelling separately for each company, we find that there exist significant company-
product relationships in our dataset. This suggests that it might be feasible for a machine-learning algorithm to relate product keywords to the companies names for the purposes of the
stock return prediction.

\subsection{Financial data analysis}
We now present our analysis of the financial data. We first study the descriptive statistics of the financial data. We tabulate the statistics in Table~\ref{tab:returns}. We note that the seven-day return has the greatest variation of all four labels - it contains the most negative return ($-20.4\%$) and the most positive return ($+26.7\%$) in the dataset. Volume, which is the amount of stock
that was traded on a given day, has a very large range in our dataset - from $1$ to $314$ million. The mean $10$ day and $30$ day volatilities are similar, but according to expectations, the $10D$ volatility has a larger standard deviation, and the difference between the minimum and maximum values is larger.

\section{Experiments}
In this section, we present preliminary experiments for benchmarking on our dataset. Our major emphasis in the experiments involve these two research questions:
\begin{itemize}
    \item Does the semantic information in tweets improve the benchmark accuracy of one-, two-,
three-, and seven-day stock return predictions?
\item Do multiple kinds of features fused together in the multi-view learning process improve
the accuracy of one-, two-, three-, and seven-day stock return predictions?
\end{itemize}

Towards this end, we use standard machine learning algorithms using the automated machine learning toolkit. We then proceed to design three benchmarks. Benchmark 1
contains only financial information. Benchmark 2 contains financial information and ``counts",
e.g. the number of news articles released, or a number of tweets released about a given stock.
Lastly, Benchmark 3 contains sentiment information from a closed resource\footnote{We obtain financial and sentiment information from Bloomberg, however, this information is not released in our dataset due to distribution restrictions}. In order to answer the
first research question, we compare Benchmarks 1-2 to Benchmarks 3, as well
as single-view and multi-view models. To answer research question two, we must compare
the results of our multi-view model with Benchmark 3, and single-view feature models.

\subsection{Features for Benchmarks}
Our features for benchmarks include a) Financial information this includes information related to trading volume, $10$ and $30$ day volatility and last price; b) Financial information and counts, which other than financial information also includes information related to news heat publication daily average, number of news articles published and number of tweets; c) Financial information, counts and sentiment information which also includes sentiment information from a proprietary algorithm. The output in each case is one-, two-, three- and seven-day stock return directions. The division into
three different benchmarks helps us to understand what kinds of financial, news and social
media signals prompt investors to buy or sell stocks. The first benchmark explains if purely
financial metrics are predictive of stock return. The second benchmark includes the count
data - the number of unexpected news publications about a given stock, as well as the number
of tweets about a given stock. This benchmark does not contain any information about the
content of these news articles and tweets; it is purely their quantity that is considered. Lastly, we include the sentiment information in the third benchmark. This last benchmark, which
uses a proprietary textual analysis metrics, is particularly useful for establishing whether our
textual information analysis brings useful insights, helping to predict stock returns with larger
accuracy.

\subsection{Features for Text-based Classifier}
The features that we consider for the textual only classifier includes a) Semantic information captured using a bag of words based vector, fastText and BERT based features; b) Sentiment polarity information using TextBlob\footnote{https://textblob.readthedocs.io/en/dev/} and a simple LSTM based sentiment prediction algorithm.
 
\subsection{Machine Learning algorithms under consideration}
We consider an out of the box implementation of automated machine learning toolbox~\cite{NIPS2015_5872}. Given that the various feature extraction methods that we use to capture different semantic information contained in the text. For example, CountVectorizer captures the information about the occurrence of a given word in a document, but it does not capture the contextual information. This can be captured by means of other algorithms, e.g. BERT. On the other hand, FastText vectors
can help us retain the information contained in rare words by making use of character-level
information. We also use an out of the box implementation of canonical correlation analysis based algorithm for multi-view learning approach. We only use CCA to obtain a highly correlated projection of vectors in each view and concatenate the projected vectors to be processed with the automated machine learning toolkit.

\subsection{Results}
\begin{table}[t]
    \centering
    \begin{tabular}{lcccc}
    \toprule
         Type & 1day & 2day & 3day & 7day \\
         \midrule
         \multicolumn{5}{c}{Single View}\\
         \midrule
         Benchmark 1 & 0.50 & 0.57 & 0.62 & 0.50 \\
         Benchmark 2 & 0.55 & 0.56 & 0.58 & 0.50 \\
         Benchmark 3 & 0.50 & 0.55 & 0.60 & 0.50 \\
         BoW & 0.50 & 0.55 & 0.57 & 0.53 \\
         fastText & 0.53 & 0.53 & 0.55 & 0.53 \\
         BERT & 0.53 & 0.52 & 0.53 & 0.53 \\
         \midrule
         \multicolumn{5}{c}{Multi View}\\
         \midrule
         BoW$+$fastText & 0.53 & 0.54 & 0.61 & 0.64 \\
         BoW$+$BERT & 0.53 & 0.57 & 0.60 & 0.54 \\
         fastText$+$BERT & 0.56 & 0.52 & 0.60 & 0.55 \\
    \bottomrule
    \end{tabular}
    \caption{Accuracy of prediction with auto-sklearn with different information content }
    \label{tab:my_label}
\end{table}
For our prediction, we test 9 different input combinations (3 benchmarks, 3 single-view and
3 multi-view vectors) with 4 labels (one-, two-, three-, and four-day returns) on 2 different
datasets (full and reduced one). For each prediction, we use 80:20 split between training and
test datasets.
We analyse the results on the full dataset. We observe that in general multiview based feature concatenation has yields better predictive performance at 3 Day return.

\subsection{Discussion}
These findings are in accordance with findings reported by~\cite{li2017discovering,zhang2018improving} on similar datasets. \cite{li2017discovering} predicts the direction of returns for a particular company, however, they do not fuse multiple data sources, taking only Twitter sentiment into account. They report the best accuracy, 66.48\%, to be with a three-day return label. However, their study also indicates that results grouped by industries are better indicators. Our dataset can be used to investigate this hypothesis further. However, a direct comparison between our results and the findings reported by these studies cannot be made. We quote these findings to illustrate that our results are in a similar range, but since we use a different dataset, the conclusions from their experiments are not strictly comparable to our study.

\section{Conclusion} In this paper, we present a labelled dataset which allows for company level tweet analysis for one-, two-, three- and seven-day stock return prediction. We also benchmark our dataset using standard machine learning approaches and show that a multi-view feature learning-based method shows promising results. Our dataset is particularly well-suited for building models for long-term, fundamental investing. Our dataset, scripts for reproducing the data and results are publically available at \url{https://github.com/ImperialNLP/stockreturnpred}.



\section{Bibliographical References}\label{reference}
\bibliographystyle{lrec}
\bibliography{refs}

\end{document}